\documentclass[a4paper]{article}

\usepackage[english]{babel}
\usepackage[utf8x]{inputenc}
\usepackage[T1]{fontenc}

\usepackage{authblk}

\usepackage[lofdepth,lotdepth]{subfig}

\usepackage{floatrow}
\newfloatcommand{capbtabbox}{table}[][\FBwidth]

\usepackage[a4paper,top=3cm,bottom=2cm,left=3cm,right=3cm,marginparwidth=1.75cm]{geometry}

\usepackage{amsmath}
\usepackage{graphicx}
\usepackage[colorinlistoftodos]{todonotes}
\usepackage[colorlinks=true, allcolors=blue]{hyperref}
\usepackage{soul}

\newcommand{\toolname}{G2D}
\newcommand{\projectpage}{\url{https://goo.gl/SS7fS6}}

\title{\toolname: from GTA to Data}

\author{Anh-Dzung Doan\thanks{Corresponding Author: dung.doan@adelaide.edu.au}, Abdul Mohsi Jawaid\thanks{abdulmohsi.jawaid@student.adelaide.edu.au}, Thanh-Toan Do\thanks{thanh-toan.do@adelaide.edu.au}, and Tat-Jun Chin\thanks{tat-jun.chin@adelaide.edu.au}}
\affil{The University of Adelaide}
\affil{Adelaide, South Australia, 5005, Australia}

\date{}

\begin{document}
\maketitle

\begin{abstract}

This document describes \toolname, a software that enables capturing videos from Grand Theft Auto V (GTA V), a popular role playing game set in an expansive virtual city. The target users of our software are computer vision researchers who wish to collect hyper-realistic computer-generated imagery of a city from the street level, under controlled 6DOF camera poses and varying environmental conditions (weather, season, time of day, traffic density, etc.).

\toolname~accesses/calls the native functions of the game; hence users can directly interact with \toolname~while playing the game. Specifically, \toolname~enables users to manipulate conditions of the virtual environment on the fly, while the gameplay camera is set to automatically retrace a predetermined 6DOF camera pose trajectory within the game coordinate system. Concurrently, automatic screen capture is executed while the virtual environment is being explored. \toolname~and its source code are publicly available at \projectpage. 

In addition, we demonstrate an application of \toolname~to generate a large-scale dataset with groundtruth camera poses for testing structure-from-motion (SfM) algorithms. The dataset and generated 3D point clouds are also made available at \url{https://goo.gl/DNzxHx}.

\end{abstract}

\section{Introduction}
\begin{figure}[h]
    \centering
    \mbox
    {
      \subfloat[][Clear]{
      \includegraphics[width=0.3\textwidth]{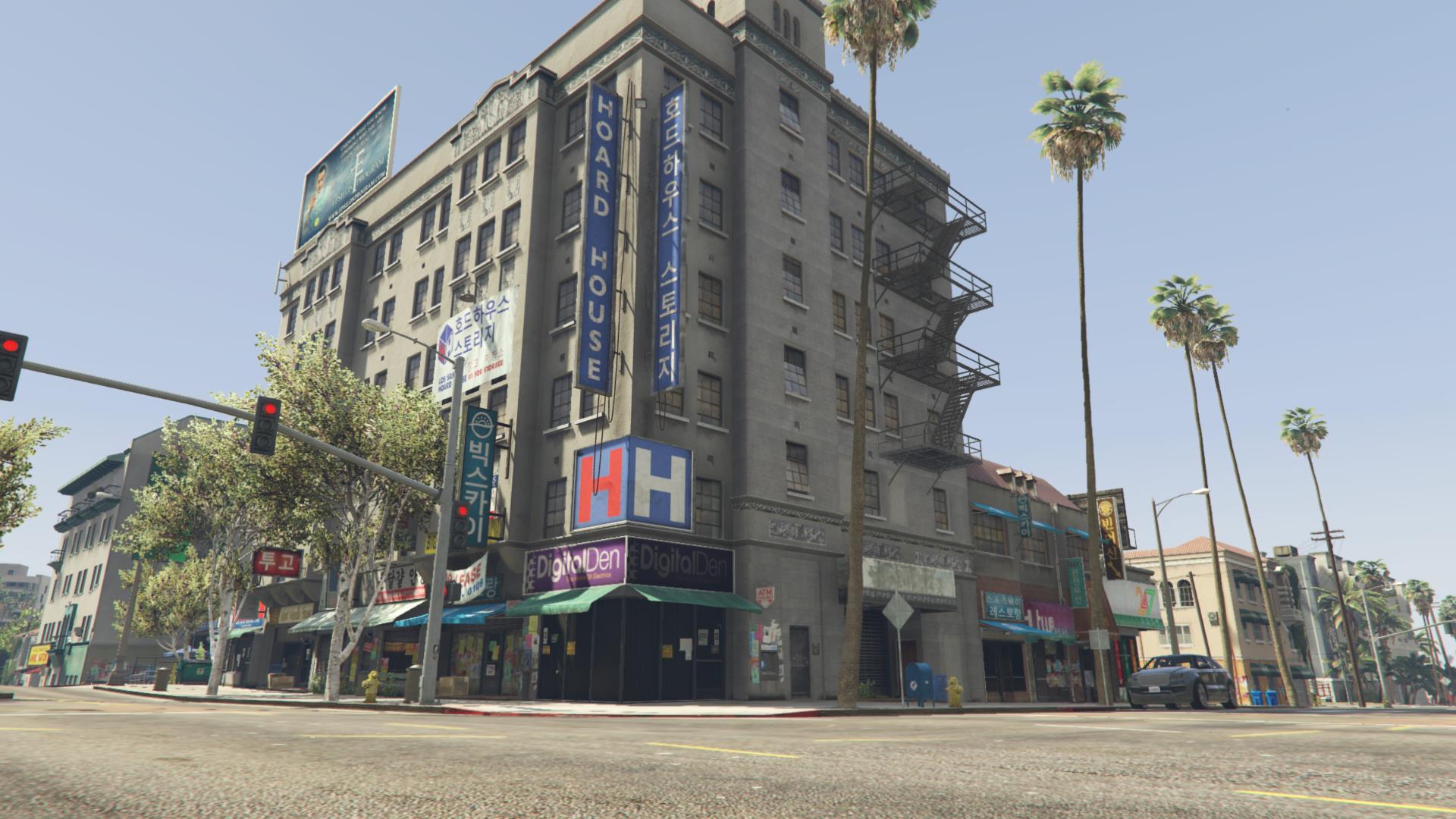}
      \label{fig:data_day-clear}}

      \subfloat[][Rain]{
      \includegraphics[width=0.3\textwidth]{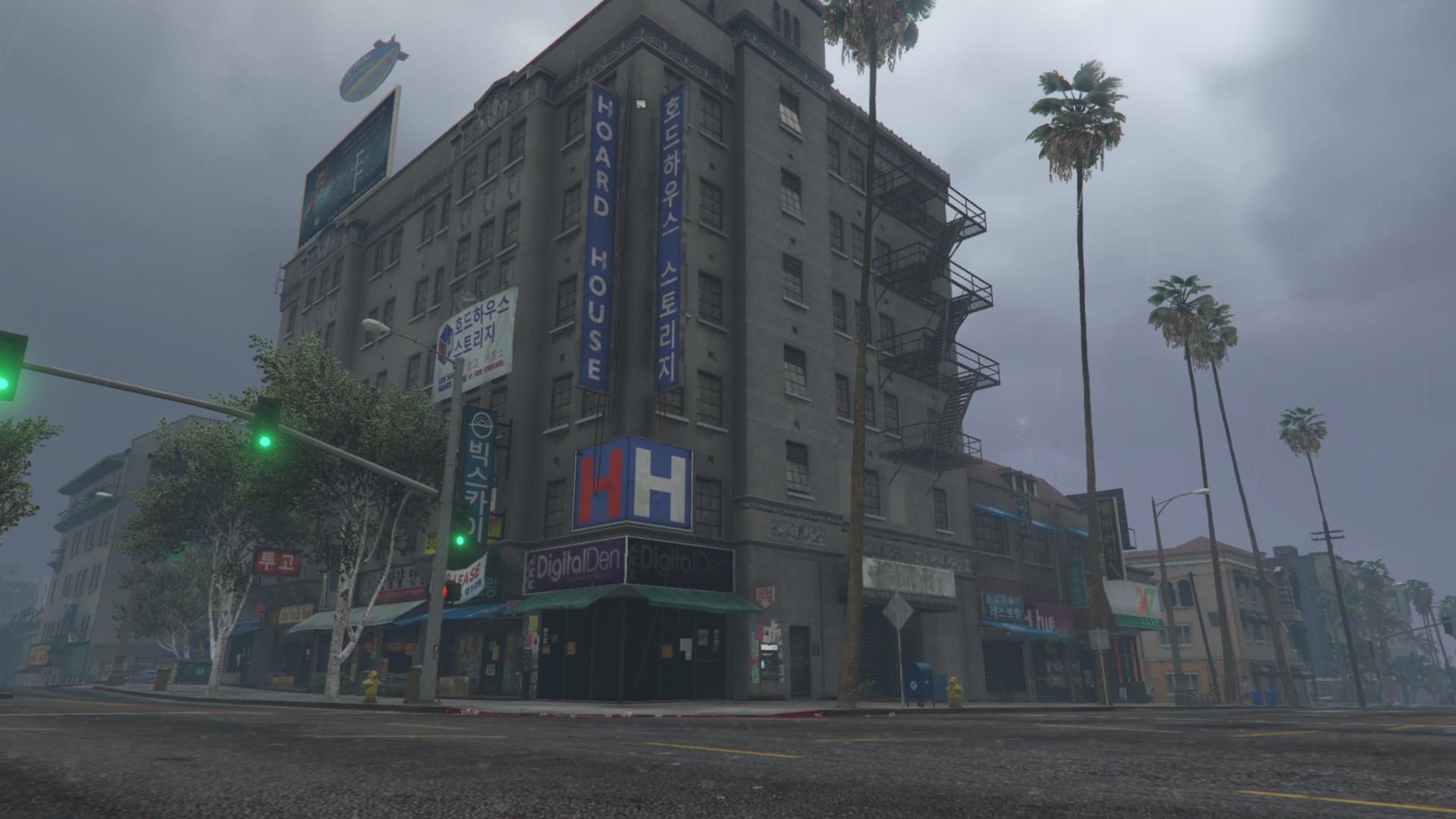}
      \label{fig:data_day-rain}}

      \subfloat[][Snow]{
      \includegraphics[width=0.3\textwidth]{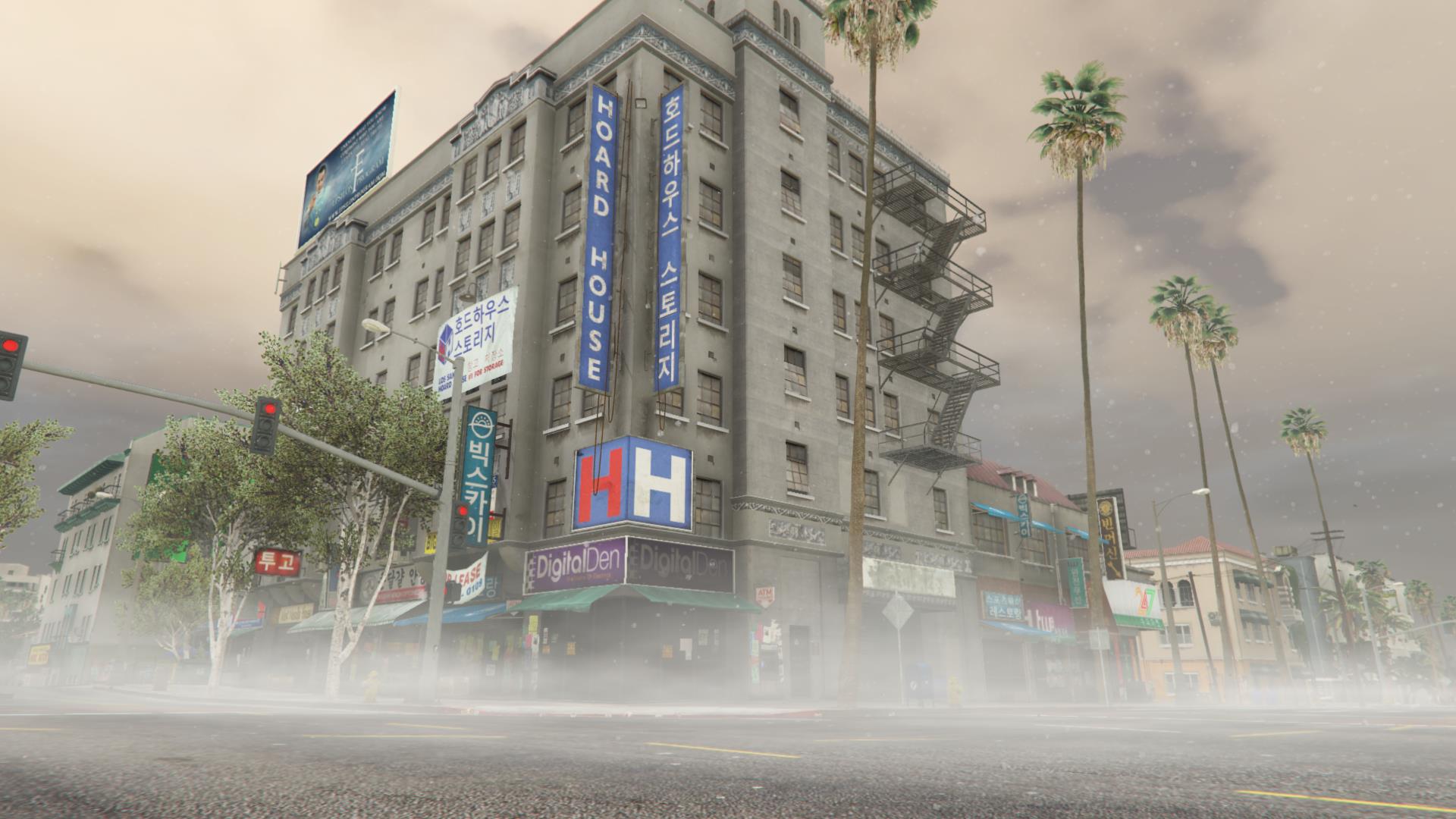}
      \label{fig:data_day-snow}}
   	}
    
  	\mbox
    {
      \subfloat[][Clear]{
      \includegraphics[width=0.3\textwidth]{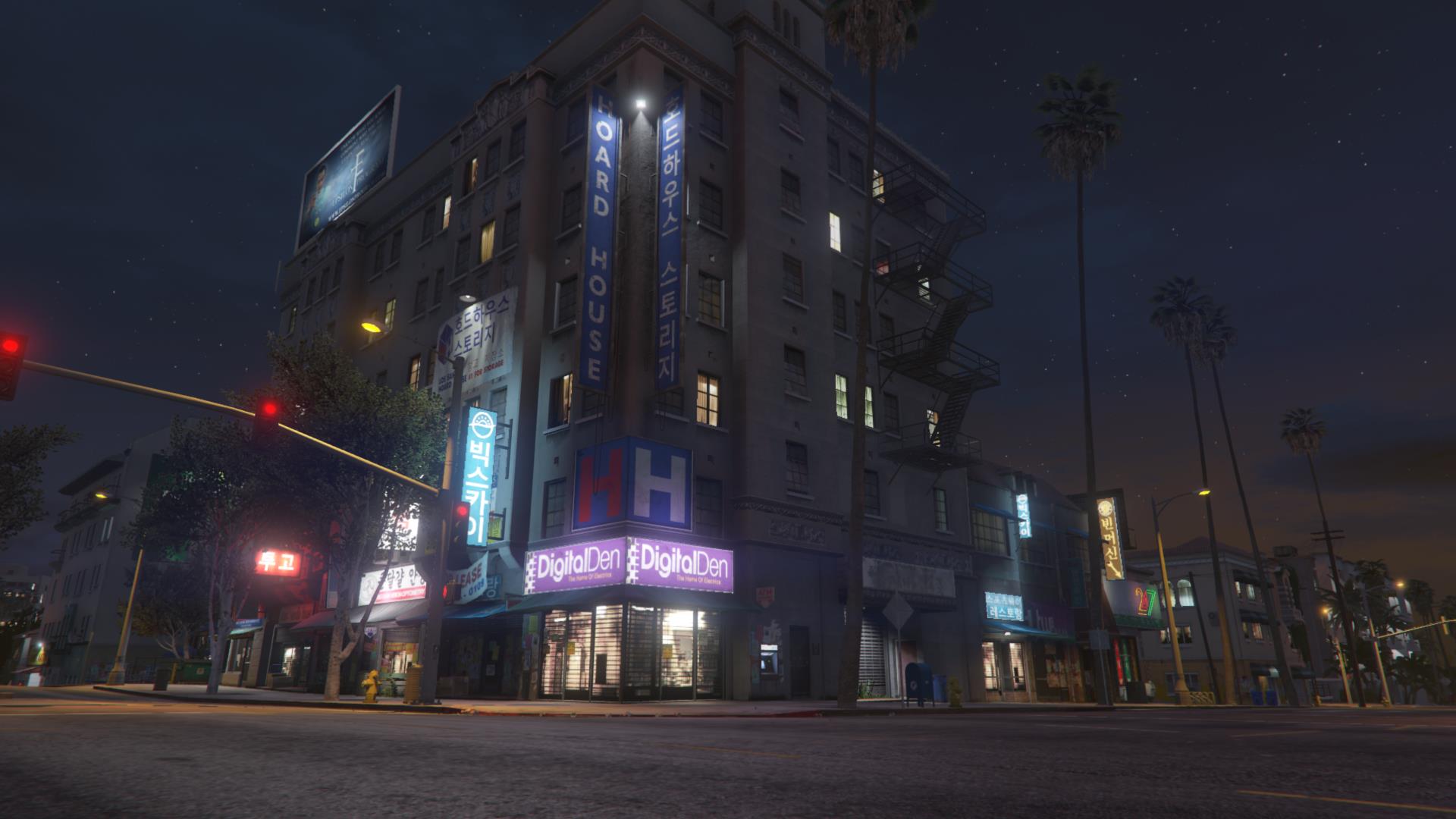}
      \label{fig:data_night-clear}}
      
      \subfloat[][Rain]{
      \includegraphics[width=0.3\textwidth]{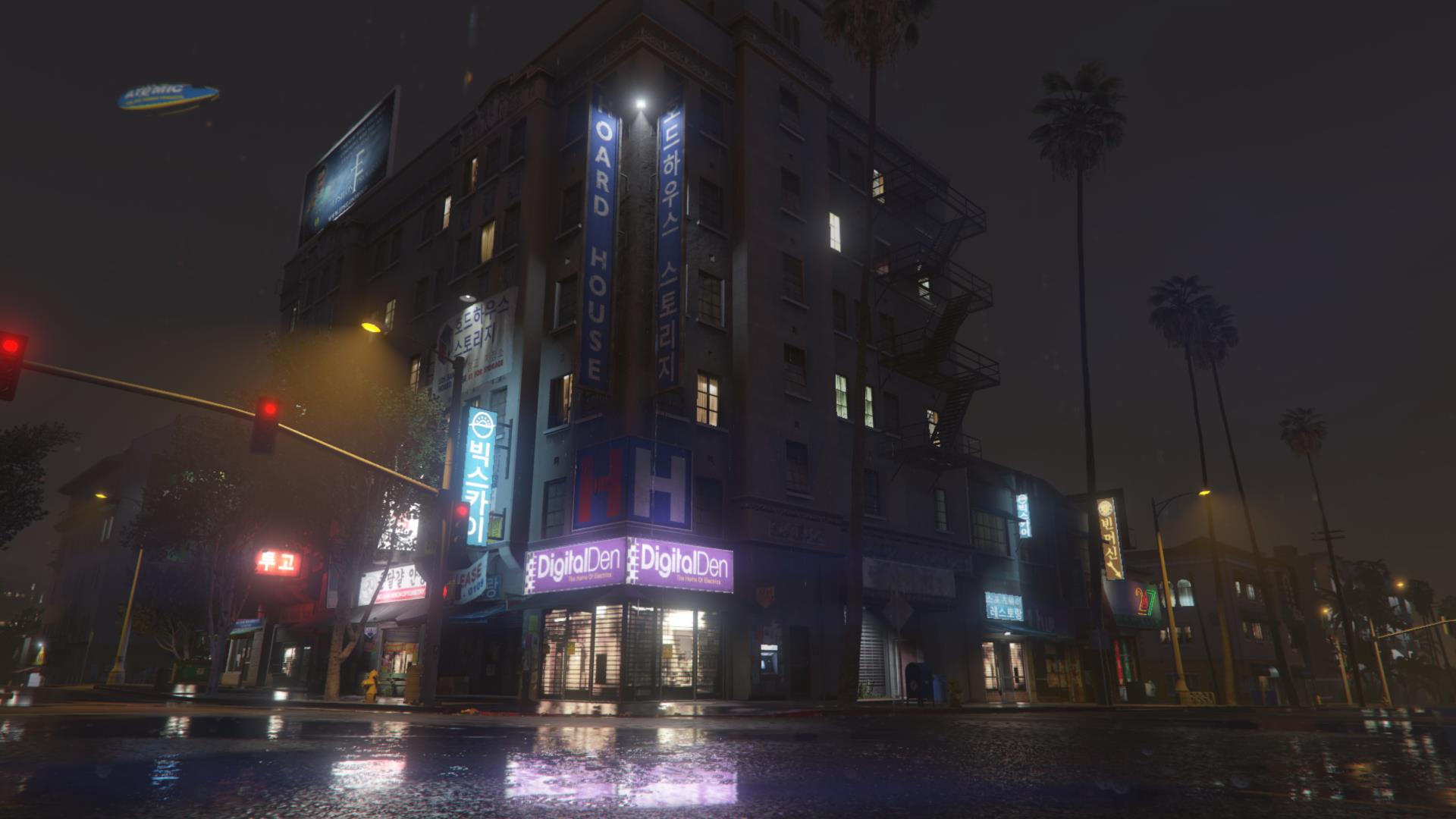}
      \label{fig:data_night-rain}}
      
      \subfloat[][Snow]{
      \includegraphics[width=0.3\textwidth]{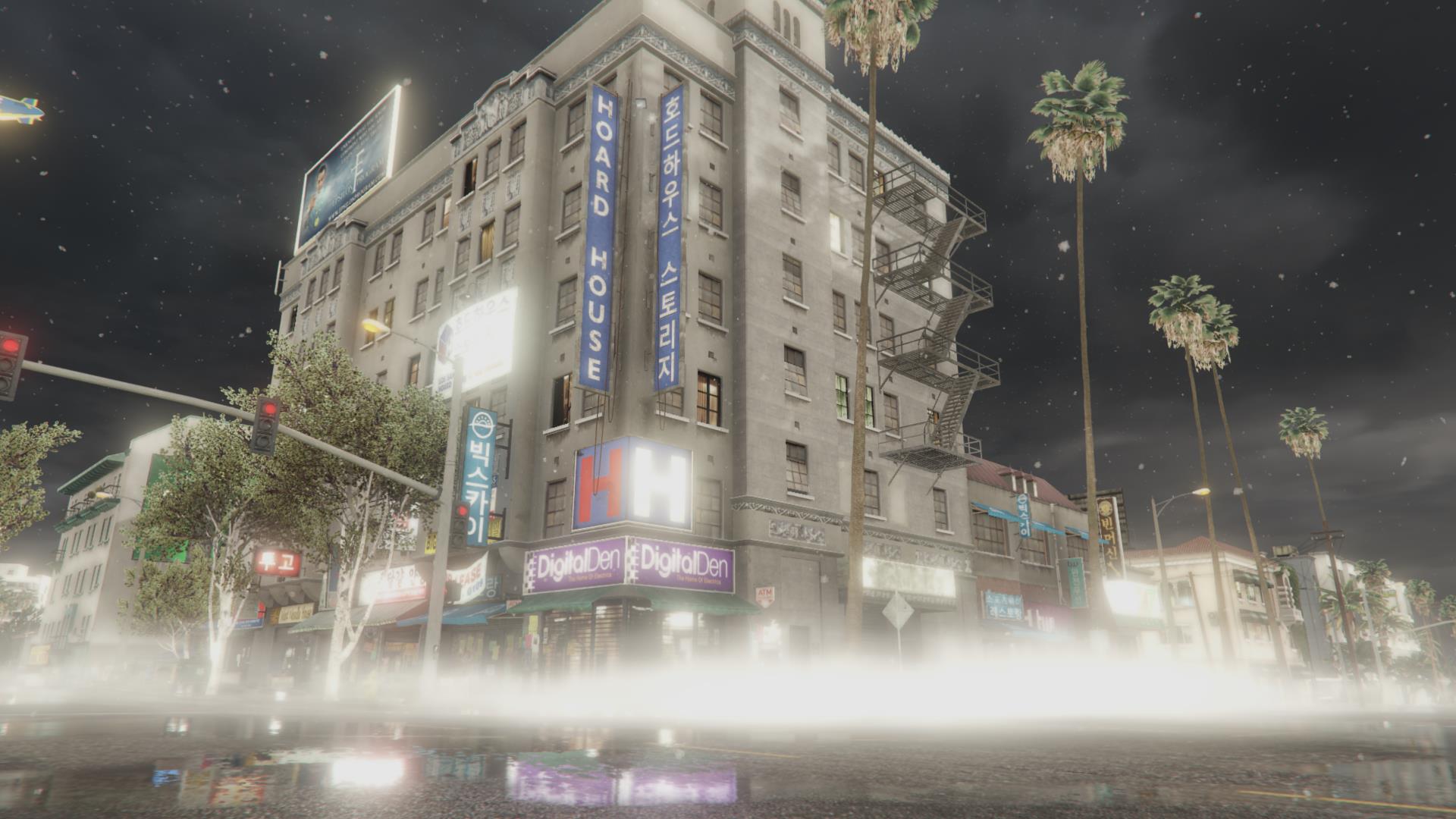}
      \label{fig:data_night-snow}}
    }

    \caption{Images under different conditions with the same camera pose. The first and second rows respectively correspond to day (12:00) and night (23:00) time.}
    \label{fig:data_example}
\end{figure}

The testing of a broad range of computer vision algorithms require datasets with accurate 6DOF groundtruth poses, e.g., SfM (structure from motion) \cite{snavely2006bundler}, visual SLAM (simultaneous localisation and mapping) \cite{davison2007monoslam}, camera pose estimation \cite{lepetit2009epnp}. Furthermore, to thoroughly assess the performance of the algorithms on realistic operating conditions, it is vital to use datasets captured under varying conditions. Indeed, recent studies~\cite{schonberger2017evaluation-local-features,zhou2016evaluation-local-features} show that the efficacy of local feature detectors, which form the first input to many computer vision algorithms, reduces significantly when presented with the same scenes under different environmental conditions (weather, time-of-day, traffic density, etc.).

Unfortunately, collecting the necessary image datasets under varying conditions is extremely costly and time-consuming. This has been a persistent obstacle towards developing and testing computer vision algorithms that are robust and reliable under varying operating conditions.

To contribute towards solving this problem, we present \toolname, an image simulator software that exploits the detailed virtual environment in GTA V. \toolname~allows users to collect hyper-realistic computer-generated imagery of an urban scene, under controlled 6DOF camera poses and varying environmental conditions (weather, season, time of day, traffic density, etc.). Users directly interact with \toolname~while playing the game; specifically, users can manipulate conditions of the virtual environment on the fly, while the gameplay camera is set to automatically retrace a predetermined 6DOF camera pose trajectory within the game coordinate system. Concurrently, automatic screen capture is executed while the virtual environment is being explored.

The output of \toolname~is a set of images with 6DOF groundtruth camera pose, captured under varying conditions; see Figure \ref{fig:data_example}.

%a tool that allowing researchers to collect datasets from Grand Theft Auto V (GTA V) which is one of the computer games with the most realistic graphics. Specifically, due to the ability of accessing to the native functions of GTA V, our tool enables users to easily manipulate the time, weather and traffic density in the game. More importantly, an exact 6DOF camera information within the game coordinate is also attached in every image collected by our tool. In other words, GTAVision enables users to collect several image sets with a consistent camera pose set under varying conditions,

\begin{figure}
	\centering
    \includegraphics[scale=0.4]{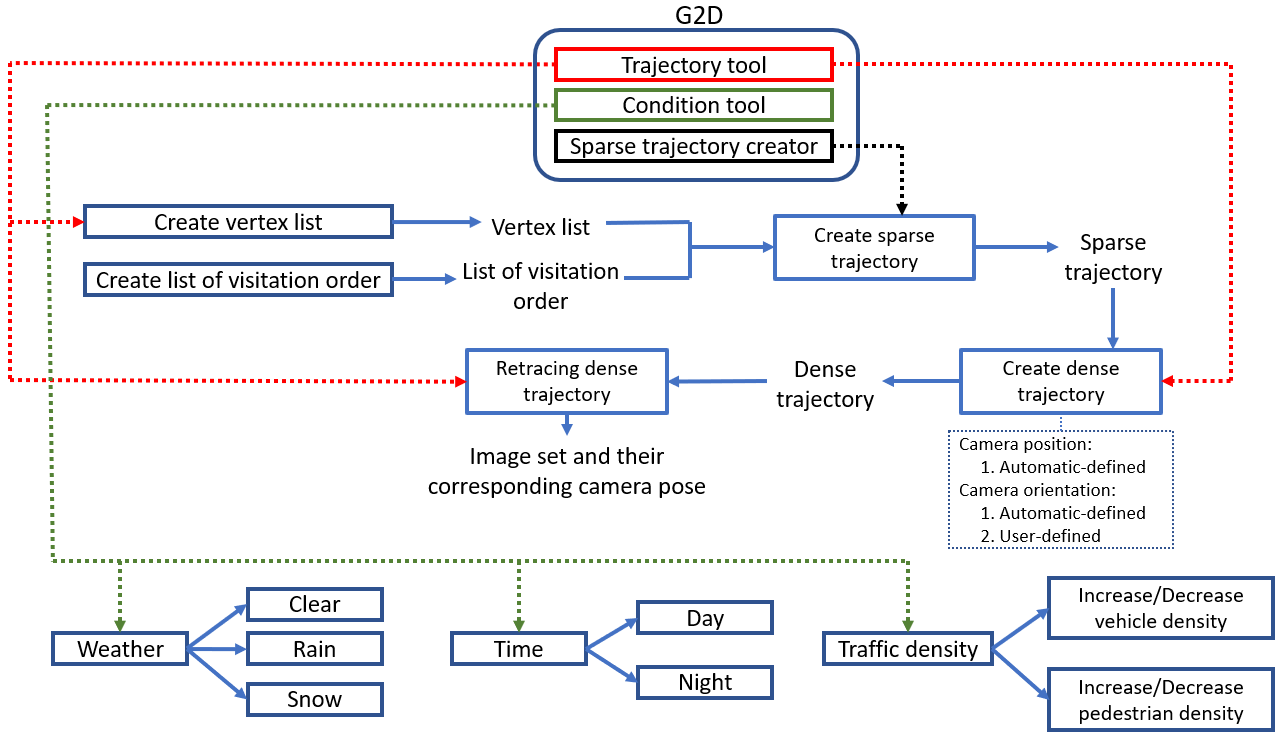}
    \caption{An illustration of all functions of \toolname. \toolname~contains 3 separate tools: trajectory tool, condition tool and sparse trajectory creator. Specifically, due to the ability of accessing to the native functions of GTA V, the condition tool enables users to easily manipulate the time, weather and traffic density. In addition, trajectory tool and sparse trajectory creator assist users to create a dense trajectory and finally collect a set of images with their corresponding camera poses. It is worth to note that because condition tool and trajectory tool are separate tools, \toolname~enables users to collect several image sets with a consistent camera pose set under varying conditions.}
    \label{fig:system_flow}
\end{figure}
An intuitive summary of the functionality of \toolname~is given in Figure \ref{fig:system_flow}. Overall, creating sparse trajectory is a user-defined step with the assistance of trajectory tool. After obtaining a sparse trajectory, a dense trajectory could be constructed in two manners: automatic-defined or user-defined orientations. The option of user-defined orientation gives a permission to define the image appearance as users preference. With the obtained dense trajectory, users could retrace that one and simultaneously collect the image set with its camera pose set. Apart from trajectory tool, condition tool could change the conditions of the game, i.e., weather, time and traffic density. Because condition tool and trajectory tool work independently, \textbf{several image sets} with a \textbf{consistent camera pose set} under \textbf{different conditions} could be collected. More details are provided in the rest of this document, as well as in the project page: \projectpage.

\section{Related Works}
There are several existing works that use virtual worlds for generating image datasets.

CARLA~\cite{carla17} provides a virtual world for autonomous driving systems. Europilot~\cite{europilot} leverages Euro Truck Simulator 2 to simulate every aspects in a driving system, e.g., wheel, brake, paddle, etc., in addition to rendering the scene. It has been suggested to use the data simulated using CARLA and Europilot to conduct experiments for autonomous-driving algorithms.

For semantic segmentation, SYNTHIA~\cite{synthia2016} is a large-scale synthetic image generator based on a virtual world constructed via the Unity  platform \cite{unity}. In that work, the synthetic images are used along with realistic images for training a deep network to improve semantic segmentation accuracy. Similarly, UnrealCV~\cite{unrealcv2017} leverages the power of Unreal engine \cite{unrealengine}, a well-known game development platform, to create synthetic images for various tasks of computer vision, based on the virtual worlds \cite{unreal_model_zoo} originally adapted from Unreal marketplace \cite{unreal_marketplace}.

Another paradigm for simulating images is to exploit readily made virtual worlds in computer games. A primary target has been GTA V, which is set in a hyper-realistic urban environment. In fact, excellent graphics is one of the reasons why GTA V is considered as the most profitable game of all time~\cite{gtav-profitable}.

Richter et al.~\cite{richter2016playing,richter2017playing} proposed a method to generate images from GTA V. Specifically, they take advantage of the communication mechanism between the game and graphics hardware by injecting a middleware between those layers, with the middleware collecting the desired game information, e.g., geometric meshes, texture maps and shaders. Based on those resources, they could then construct the groundtruth for a variety of computer vision problems, including visual odometry. In fact, the camera poses are obtained through the recovery from the recorded meshes and their transformation matrices..

Different from Richter et al., our software \toolname~accesses the native functions of GTA V. This enables us to directly attain the camera pose in every frame of the game. Additionally, users can control the environmental conditions through functions developed in \toolname.

\section{Methodology}

\subsection{Scripthook V}
\toolname~is based on Scripthook V \cite{ScripthookWebsite}, a library developed by Alexander Blade that provides access to the native functions of GTA V (the usage Scripthook V distinguishes our \toolname~from Richter et al.). The original aim of Scripthook V is to provide a framework to construct modifications (``mods") to the game. Currently, a wide range of fascinating mods are available \cite{GTAmods}, e.g., Invisibility Cloak \cite{invi_mod_gta} that can make the protagonist invisible. The list of native functions supported by Scripthook V could be found on \cite{ScripthookFunctionList}.

\subsection{Constructing Trajectory}
\toolname~defines two types of camera trajectories: \emph{sparse} and \emph{dense} trajectories. A sparse trajectory consists of a set of vertices (a set of positions on the ``top down" 2D map of the virtual environment), and an order in which to visit the vertices. Users specify sparse trajectories. Then, given a sparse trajectory, a dense trajectory is generated automatically by \toolname. Basically, the tool traces a continuous path along the dense trajectory and captures the scene as observed from the gameplay camera at \textbf{60 frames per second}\footnote{We measure this performance from a workstation Intel(R) Core(TM) i7-6700 @ 3.40GHz, RAM 16GB, NVIDIA GeForce GTX 1080 Ti and the maximum graphical configuration for GTA V}, along with the 6DOF camera pose at each frame. In other words, \toolname~still guarantees the normal operation of the game as well as the collected dataset is in the standard video rate.

Figure~\ref{fig:trajectory_example} shows an example, while more details are available in the following.

\subsubsection{Sparse Trajectory (user-defined)}
\begin{figure}
	\centering
	\includegraphics[scale=0.4]{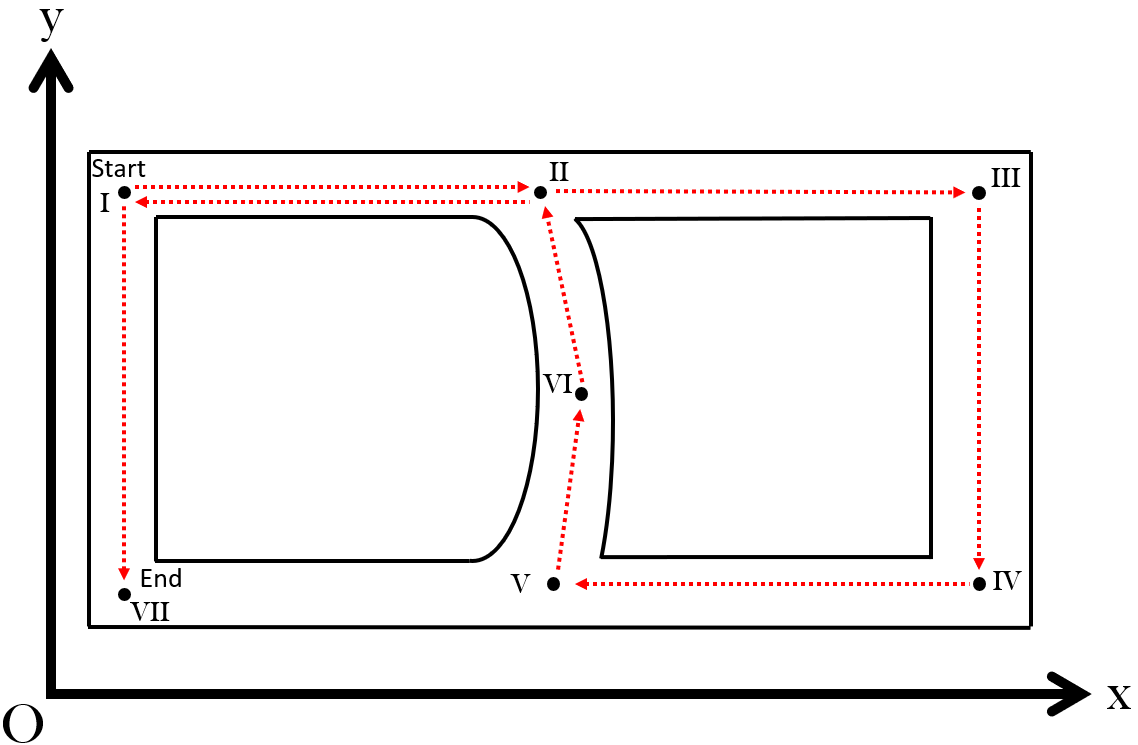}
    \caption{Illustrating sparse and dense trajectories on a 2D map/coordinate system. Points labelled with Roman numerals indicate the vertices of a sparse trajectory; the order of visitation of the vertices is given by the red arrows. The sampled frames along with the 6DOF camera pose at each frame while automatically traversing the sparse trajectory gives rise to a dense trajectory.}
    \label{fig:trajectory_example}
\end{figure}

The vertices $\{I, II, \dots\}$ of a sparse trajectory are defined by their coordinates on a 2D map $\{ (x_{I}, y_{I}), (x_{II}, y_{II}), \dots \}$. The order of visitation is specified using an index
\begin{align*}
ORDER_{I} &= \{a,b,c,\dots\}\\
ORDER_{II} &=\{x,y,z,\dots\}\\
&\vdots
\end{align*}
where $a,b,c$ are integers indicating the order in which vertex $I$ is visited in the desired sequence (similarly $x,y,z$ for vertex $II$). For the example in Figure \ref{fig:trajectory_example}, the index would be 
\begin{align*}
ORDER_{I} &=\begin{Bmatrix}1,8\end{Bmatrix}\\
ORDER_{II} &=\begin{Bmatrix}2,7\end{Bmatrix}\\
ORDER_{III} &=\begin{Bmatrix}3\end{Bmatrix}\\ 
ORDER_{IV} &=\begin{Bmatrix}4\end{Bmatrix}\\
ORDER_{V} &=\begin{Bmatrix}5\end{Bmatrix}\\
ORDER_{VI} &=\begin{Bmatrix}6\end{Bmatrix}\\
ORDER_{VII} &=\begin{Bmatrix}9\end{Bmatrix}
\end{align*}
In \toolname, the vertex and order of visitation are specified in the files {\tt vertex.txt} and {\tt vertex\_order.txt} respectively.

%Regarding the file format, the 1st, 2nd, 3rd, etc lines within those files contains the data for vertex I, II, III, etc.

\subsubsection{Dense Trajectory (generated automatically)}
\label{sec:dense_trajectory}

Given a user-specified sparse trajectory, \toolname~moves the protagonist of the game to automatically follow the trajectory. The orientation (rotation) of the camera while the movement is being executed can be specified in two modes:
\begin{itemize}
\item First-person view mode: \toolname~attaches the gameplay camera to the ``eyes" of the protagonist, and the viewing direction always points forward without the need to handle camera orientations by the user.
\item Third-person view mode: While the protagonist automatically moves along the trajectory, the user can use the mouse to control the orientation of the camera.
\end{itemize}
While the environment is being explored, \toolname~calls the relevant native functions and performs the necessary computations to obtain the 6DOF pose of the gameplay camera at each frame. Every 6DOF pose is stored in line-by-line manner within the file {\tt trajectory\_dense.txt}. Each line within {\tt trajectory\_dense.txt} has the following format: 

\begin{center}
{\tt <protagonist position XYZ> <camera position XYZ> <camera rotation XYZ>}
\end{center}

It is worth noting that because the dense trajectory is simply an editable text file, users could easily open the dense trajectory and make some manual modifications to create a noisy version of trajectory, hence a more challenging dataset could be generated.

%\subsection{Data Sampling}
\subsection{Retracing a dense trajectory}
With the obtained dense trajectory, \toolname~opens the file {\tt trajectory\_dense.txt}, sequentially loads each line within that file and then sets 6DOF pose to the camera object. With each 6DOF value, \toolname~performs a screenshot of the screen rendered by the camera object.

\toolname~stores all image data along with their corresponding 6DOF pose within {\tt 6dpose\_list.txt} as the following format:

\begin{center}
{\tt<image file name 1> <camera position XYZ> <camera rotation XYZ>}

{\tt<image file name 2> <camera position XYZ> <camera rotation XYZ>}

.
.
.

{\tt<image file name N> <camera position XYZ> <camera rotation XYZ>}
\end{center}

All the images and their 6DOF pose are automatically and fully generated from the predetermined dense trajectory as the explanation in section \ref{sec:dense_trajectory}. Therefore, before carrying out the function of retracing the dense trajectory, users could change the environmental conditions (i.e. weathers, time and traffic density) as their preference to attain their desired dataset.

\subsection{Changing the  the environmental conditions}
 \toolname~provides the functions that could support users to change environmental conditions. There are three different settings regarding environmental conditions:

\begin{itemize}
\item Regarding the weather, \toolname~allows user to select between clear, rain or snow. 

%\toolname~uses the {\tt GAMEPLAY::SET\_OVERRIDE\_WEATHER} to manipulate the weather. That functions receives 1 parameter that indicates the desired weather. That parameter value could be: {\tt "EXTRASUNNY", "CLEAR", "CLOUDS", "SMOG", "FOGGY", "OVERCAST", "RAIN", 
%\vskip -1.5mm
%"THUNDER", "NEUTRAL", "SNOW", "BLIZZARD", "SNOWLIGHT", "XMAS"}

\item In terms of the time, \toolname~implements day time (at 12:00) and night time (at 23:00). 

%\toolname~uses {\tt TIME::SET\_CLOCK\_TIME} to control the time. That function receives 3 input parameters representing hour, minute and second.

\item With regard to the traffic density, \toolname~assists users to increase or decrease two types of traffic density, i.e. vehicle and pedestrian. The density value varying from $0$ to $1$ represents from none to normal numbers of pedestrians/vehicles on the road.

%{\tt PED::SET\_PED\_DENSITY\_MULTIPLIER\_THIS\_FRAME} and 
%\vskip -1.5mm
%{\tt VEHICLE::SET\_VEHICLE\_DENSITY\_MULTIPLIER\_THIS\_FRAME} functions are utilized to respectively handle the pedestrian and vehicle density. Those both functions receive 1 input parameter indicating the density value.
\end{itemize}

\subsection{Unit Conversion}
In the testing of some algorithms, it may be useful to conduct metric conversion of the distances in the 2D map of the game. To this end, we perform the following trick:
\begin{itemize}
\item Make the game protagonist walk in the environment, and record the positions (in the 3D game coordinate) of the protagonist at every 2-3 steps.
\item Calculate the average walking-stride length of the protagonist in unit-distances of the 3D game coordinate (roughly 0.9 units based on the 3D game coordinate that we used).
\end{itemize}
According to \cite{StrideWalkLength}, the average walking stride length for a male adult is about $0.762$ meters, hence 1 unit-distance in the 3D game coordinate is equal to about 0.85 meters.

\section{Sample application---testing SfM pipelines}
\begin{figure}
  \begin{floatrow}
    \ffigbox{%
      \includegraphics[scale=0.4]{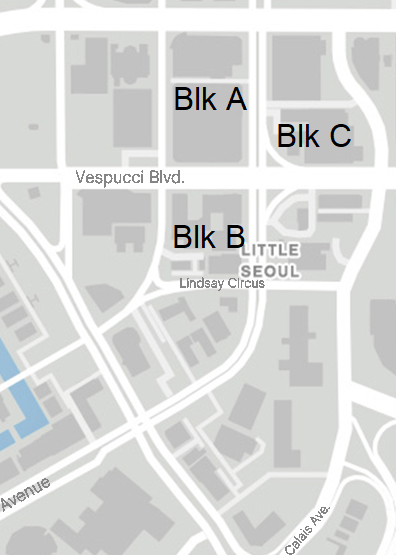}
    }{%
      \caption{Locations of our three dataset in Little Seoul region}%
      \label{fig:dataset_location_in_game}
    }
    
    \capbtabbox{%
      \begin{tabular}{|c|c|c|}
            \hline
            Name & \# images & \# images for \\
                 &			 &	reconstruction\\
            \hline
            \hline
            Blk A & 7,372 & 369\\
            \hline
            Blk B & 6,092 & 305\\
            \hline
            Blk C & 7,658 & 383\\
            \hline
        \end{tabular}
    }{%
      \caption{The dataset collected using \toolname, each dataset is a block in Little Seoul region in GTA V}%
      \label{tab:dataset_info}
    }
  \end{floatrow}
\end{figure}

In this section, we would like to use the datasets collected by \toolname~for structure from motion (SfM), which is one of the fundamental problems in geometric computer vision. In order to leverage the camera poses in game coordinate as the groundtruth, we employ the registration \cite{umeyama1991least} inside RANSAC to align the camera positions reconstructed by SfM to the game coordinate. Our experimental evaluation is quite similar to \cite{wilson2014sfm1d}, but they use the camera position reconstructed by Bundler \cite{snavely2006bundler} as groundtruth while we use the exact camera poses in the game coordinate. The structure from motion framework that we use is COLMAP \cite{schoenberger2016sfm} \cite{schoenberger2016mvs}.

\begin{figure}[h]
    \centering
    \mbox
    {
      \subfloat[][]{
      \includegraphics[width=0.2\textwidth]{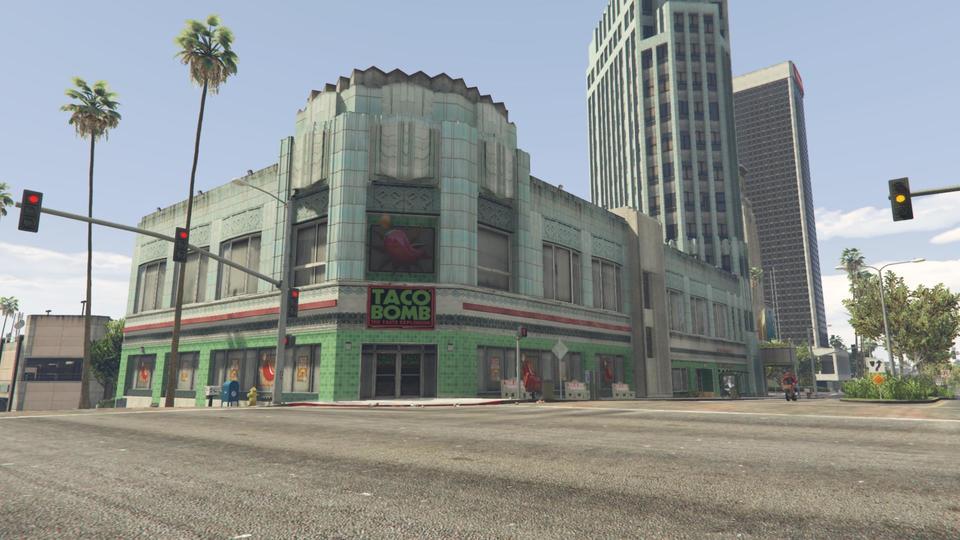}}

      \subfloat[][]{
      \includegraphics[width=0.2\textwidth]{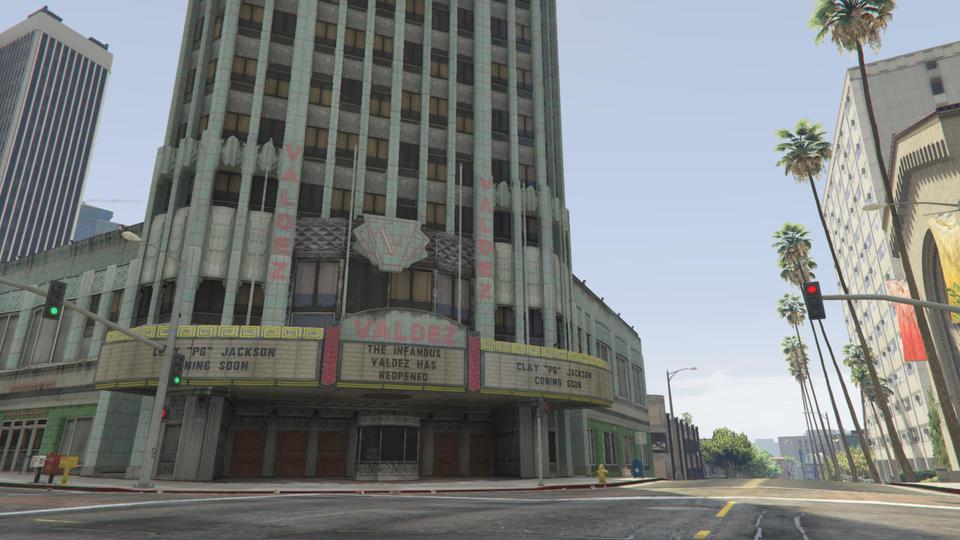}}

      \subfloat[][]{
      \includegraphics[width=0.2\textwidth]{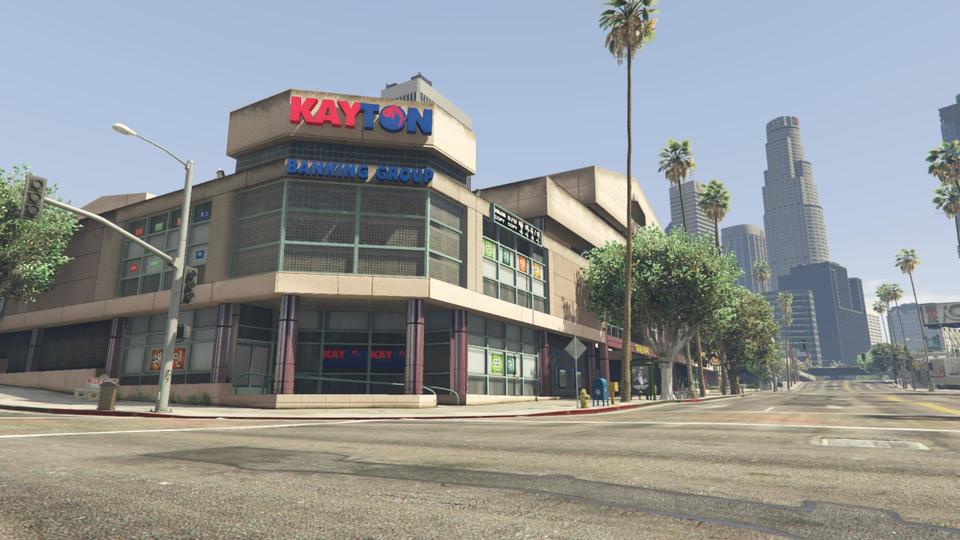}}
      
      \subfloat[][]{
      \includegraphics[width=0.2\textwidth]{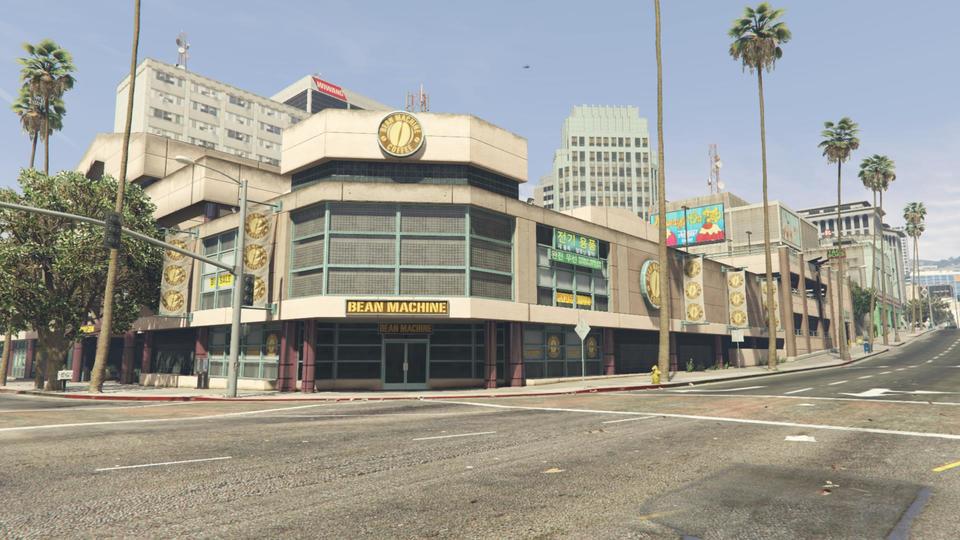}}
   	}
    
  	\mbox
    {
      \subfloat[][]{
      \includegraphics[width=0.2\textwidth]{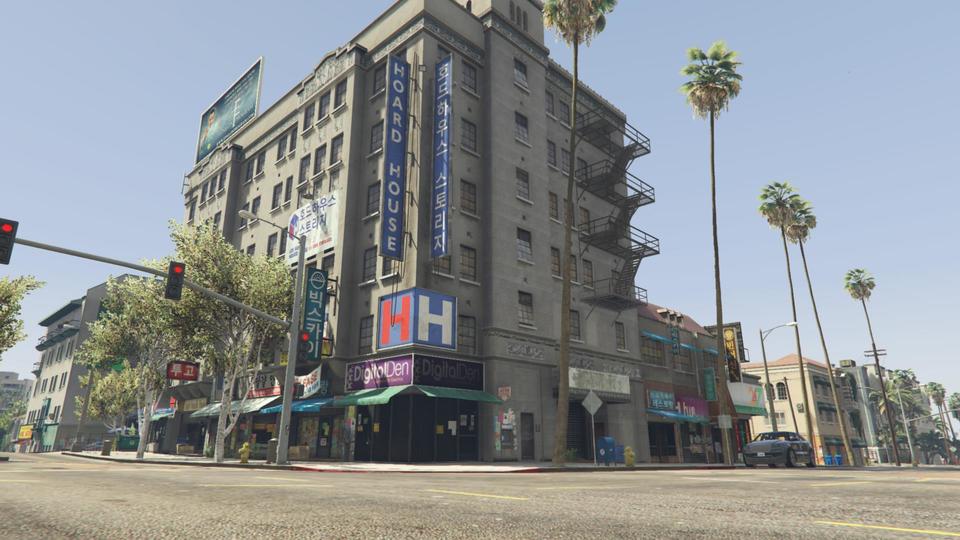}}

      \subfloat[][]{
      \includegraphics[width=0.2\textwidth]{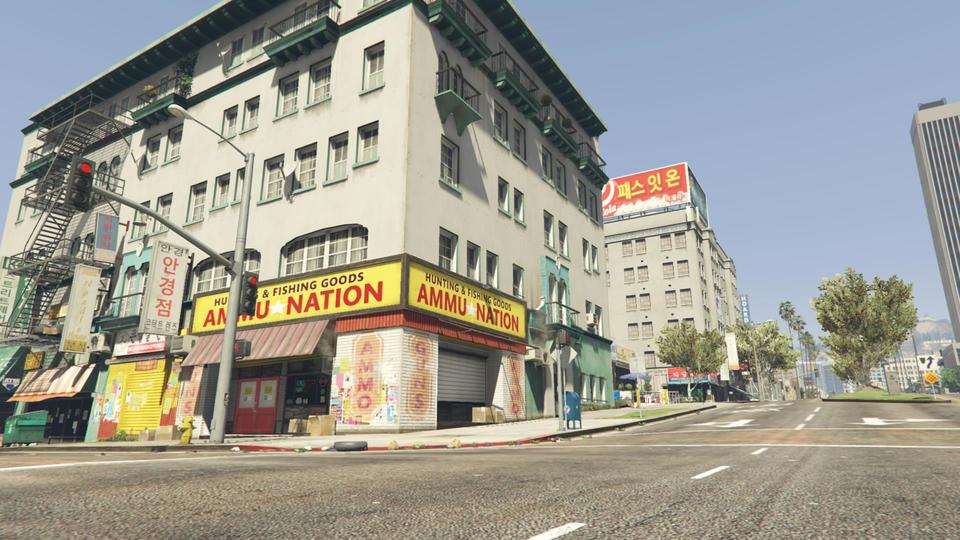}}

      \subfloat[][]{
      \includegraphics[width=0.2\textwidth]{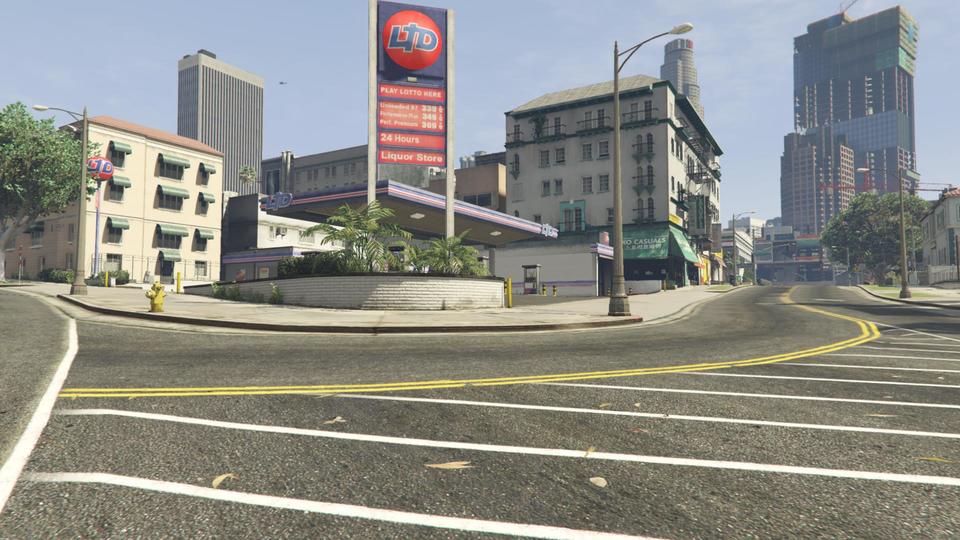}}
      
      \subfloat[][]{
      \includegraphics[width=0.2\textwidth]{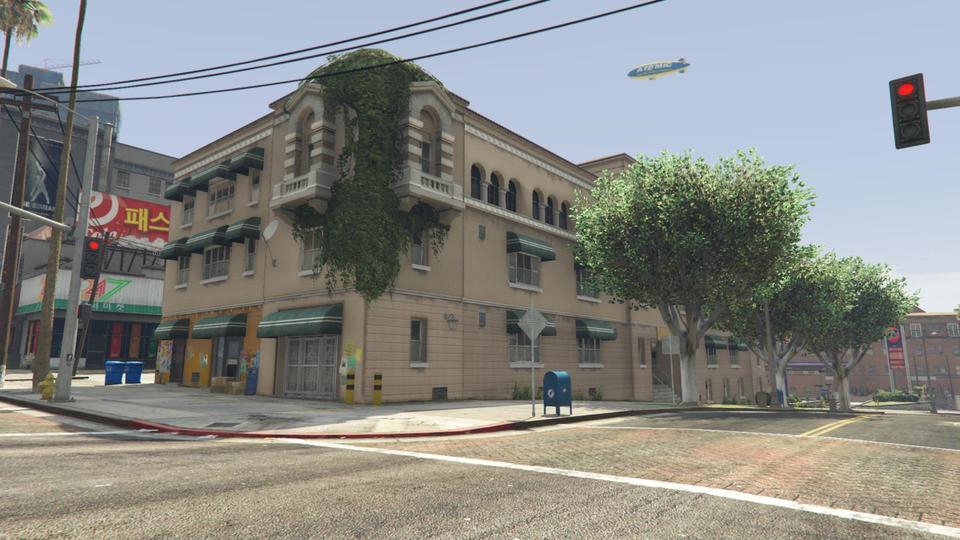}}
    }
    
    \mbox
    {
      \subfloat[][]{
      \includegraphics[width=0.2\textwidth]{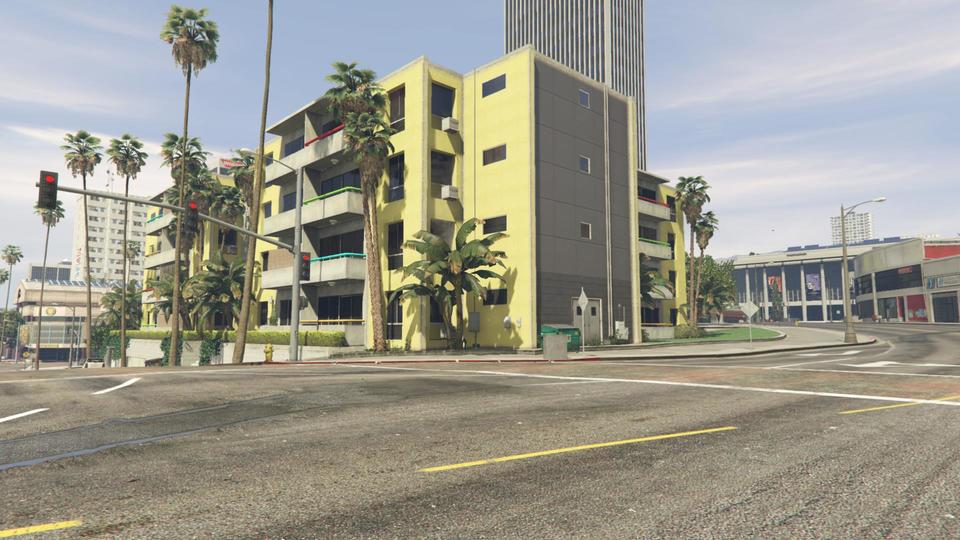}}

      \subfloat[][]{
      \includegraphics[width=0.2\textwidth]{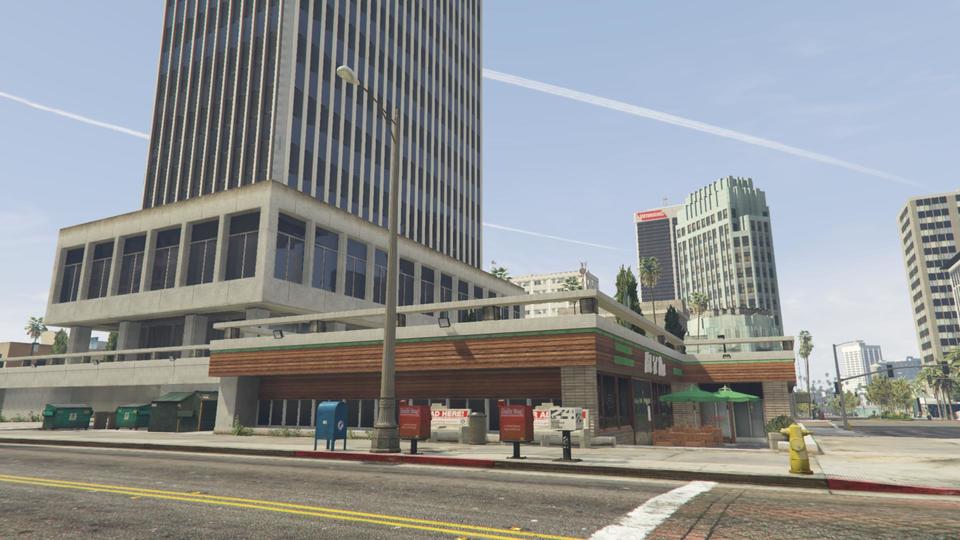}}

      \subfloat[][]{
      \includegraphics[width=0.2\textwidth]{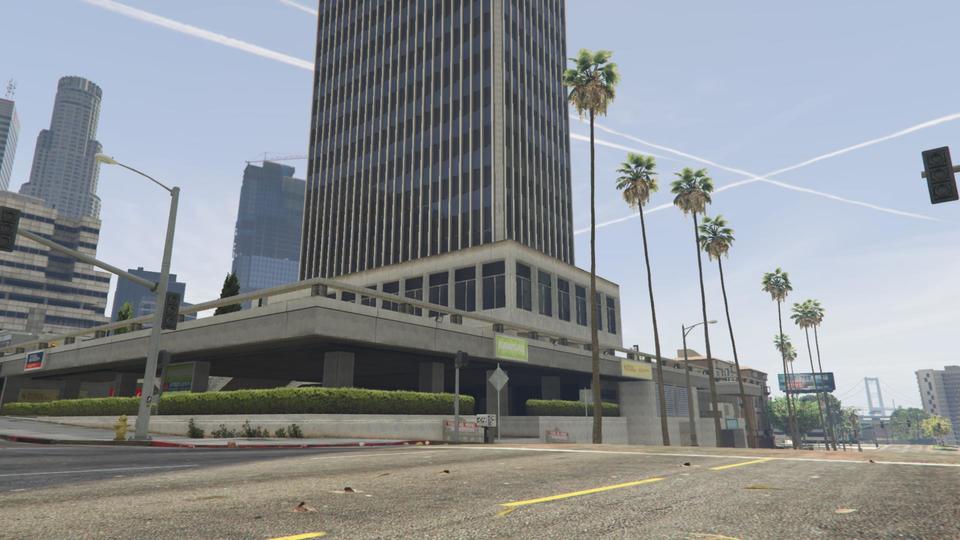}}
      
      \subfloat[][]{
      \includegraphics[width=0.2\textwidth]{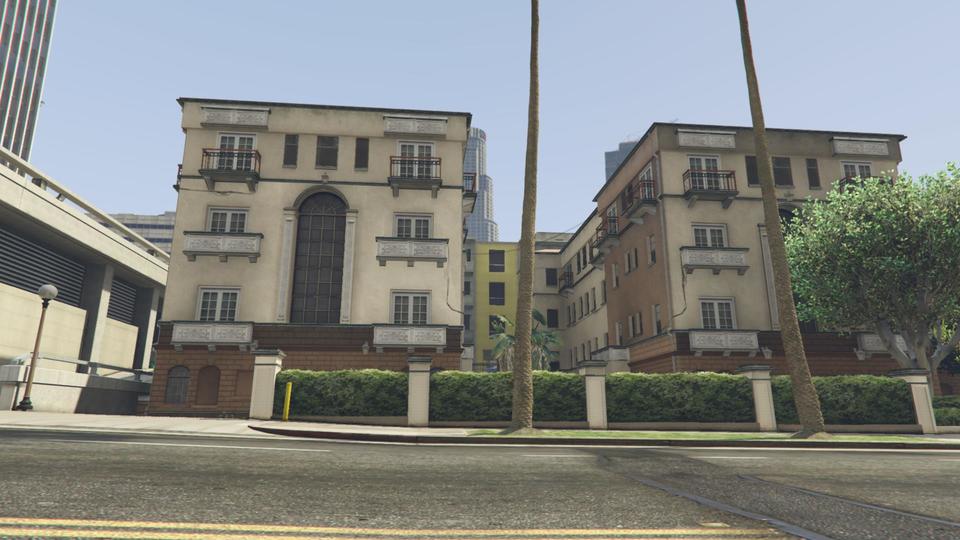}}
    }

    \caption{Some sample images in our dataset; 1st, 2nd and 3rd rows are respectively samples of Blk A, Blk B and Blk C.}
    \label{fig:dataset_sample_images}
\end{figure}

\begin{table}
	\centering
  	\begin{tabular}{|c|c|c|c|c|c|}
        \hline
        Name & size & \# registered & \# points & average error & median error \\
        	 &	    & images        &           & (m)           & (m)	       \\
        \hline
        \hline
        Blk A & 369 & 369 & 105,342 & 0.31 & 0.19\\
        \hline
        Blk B & 305 & 305 & 86,186 & 0.12 & 0.07\\
        \hline
        Blk C & 383 & 383 & 88,405 & 0.19 & 0.13\\
        \hline
  	\end{tabular}
  	\caption{The reconstruction results on our three datasets}
  	\label{tab:recon_result}
\end{table}

\begin{figure}
    \centering
    
      \subfloat[][Blk A]{
      \includegraphics[width=0.8\textwidth]{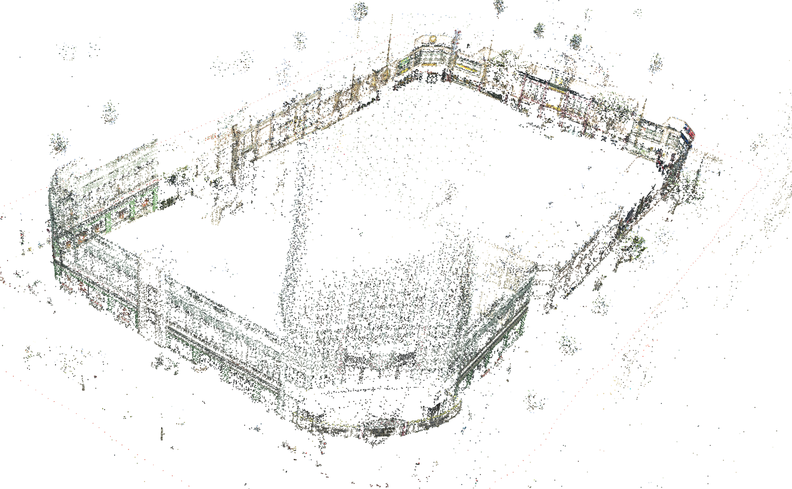}}

      \subfloat[][Blk B]{
      \includegraphics[width=0.8\textwidth]{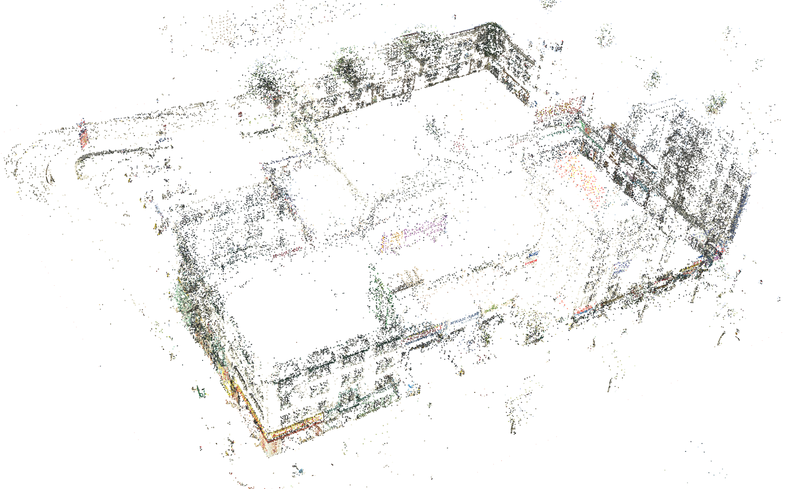}}
      
       \subfloat[][Blk C]{
      \includegraphics[width=0.8\textwidth]{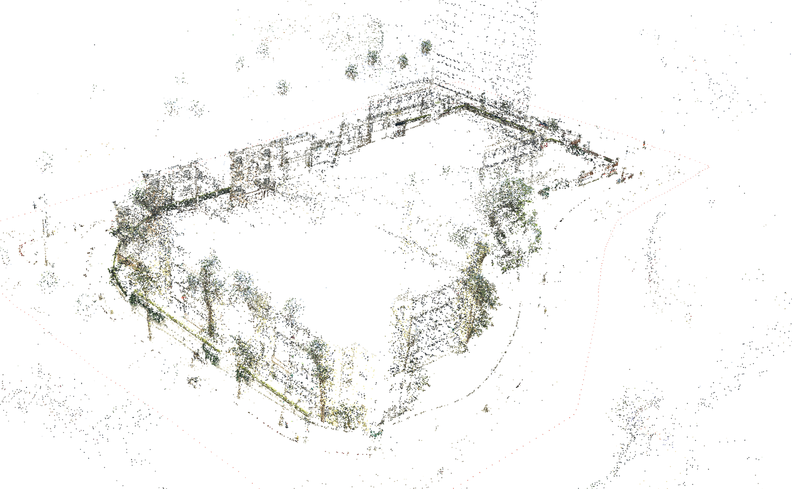}}

    \caption{Renders of 3D point cloud of our three datasets}
    \label{fig:recon_result_vis}
\end{figure}

With regards to our datasets, we collect three different datasets from three blocks in the Little Seoul region. Figure \ref{fig:dataset_location_in_game} demonstrates the location of those blocks and Table \ref{tab:dataset_info} shows that number of collected images. Actually, we only use the subset of those images for the reconstruction. Figure \ref{fig:dataset_sample_images} illustrates the sample images within our three datasets. In terms of game conditions, the time, weather and traffic density are day (12:00), clear and 0 respectively.

Table \ref{tab:recon_result} reports the reconstruction results and Figure \ref{fig:recon_result_vis} visualizes our obtained 3D point cloud.

\section{Conclusion}
This document presents \toolname~that could be utilized to collect the dataset in Grand Theft Auto V. Due to the capability of accessing to the native functions, \toolname~allows users to control the various environmental conditions within the game, i.e. weather, time and traffic density. In addition, \toolname~samples a set of images with their corresponding camera poses in the game coordinate. Apart from it, \toolname~is open-sourced, hence users could modify it as their preference to collect the desired dataset. The section of sample application demonstrates the use of those camera poses as groundtruth for the structure from motion.
\bibliographystyle{IEEEtran}
\bibliography{bibliography}

\end{document}